\documentclass[runningheads,a4paper]{llncs}

\usepackage[dvipsnames,svgnames,x11names,hyperref, table]{xcolor}
\usepackage[colorlinks = true,
            allcolors = NavyBlue
            ]{hyperref}

\usepackage[utf8]{inputenc}
\usepackage[T1]{fontenc}
\usepackage{caption}
\usepackage{multirow}

\usepackage{latexsym}

\usepackage{makecell}

\usepackage{enumitem}

\usepackage[table]{xcolor}
\usepackage{booktabs}
\usepackage{siunitx}
\makeatletter
\@ifundefined{c@rownum}{%
  \let\c@rownum\rownum
}{}
\@ifundefined{therownum}{%
  \def\therownum{\@arabic\rownum}%
}{}
\makeatother

\usepackage{etoolbox}
\robustify\bfseries

\usepackage{xspace}
\makeatletter
\DeclareRobustCommand\onedot{\futurelet\@let@token\@onedot}
\def\@onedot{\ifx\@let@token.\else.\null\fi\xspace}
\def\eg{e.g\onedot} \def\Eg{E.g\onedot}

\def\etc{etc\onedot} \def\vs{vs\onedot}
\def\wrt{w.r.t\onedot} 
\def\etal{et al\onedot}

\makeatother

\usepackage{amssymb}
\usepackage{graphicx}

\usepackage{url}

\begin{document}

\mainmatter  

\title{SART - Similarity, Analogies, and Relatedness for Tatar~Language: New Benchmark~Datasets for Word~Embeddings Evaluation}

\titlerunning{SART - Similarity, Analogies, and Relatedness for Tatar~Language}

\author{Albina Khusainova\inst{1} \and Adil Khan\inst{1} \and Adín Ramírez Rivera\inst{2}}
\authorrunning{Albina Khusainova et al.}   
%
\tocauthor{Albina Khusainova, Adil Khan, Adín Ramírez Rivera}
\institute{Innopolis University,  Universitetskaya, 1, 420500, Innopolis, Russia\\
\email{\{a.khusainova, a.khan\}@innopolis.ru},
\and
University of Campinas (UNICAMP), 13083-970, Campinas, Brazil\\
\email{adin@ic.unicamp.br}
}

\toctitle{}
\tocauthor{}
\maketitle

\begin{abstract}
 There is a huge imbalance between languages currently spoken and corresponding resources to study them.  Most of the attention naturally goes to the ``big'' languages---those which have the largest presence in terms of media and number of speakers. Other less represented languages sometimes do not even have a good quality corpus to study them. In this paper, we tackle this imbalance by presenting a new set of evaluation resources for Tatar, a language of the Turkic language family which is mainly spoken in Tatarstan Republic, Russia. 
  
 We present three datasets: Similarity and Relatedness datasets that consist of human scored word pairs and can be used to evaluate semantic models; and Analogies dataset that comprises analogy questions and allows to explore semantic, syntactic, and morphological aspects of language modeling. All three datasets build upon existing datasets for the English language and follow the same structure. However, they are not mere translations. They take into account specifics of the Tatar language and expand beyond the original datasets. We evaluate state-of-the-art word embedding models for two languages using our proposed datasets for Tatar and the original datasets for English and report our findings on performance comparison.
 
 The datasets are available at \url{https://github.com/tat-nlp/SART}
\end{abstract}

\section{Introduction}
Word embeddings have become almost an intrinsic component of NLP systems based on deep learning. Therefore, there is a need for their evaluation and comparison tools. It is not always computationally feasible to evaluate embeddings directly on the task they were built for; that is why there is a need for inexpensive preliminary evaluations. For example, there are similarity/relatedness tests, where human judgements are obtained for pairs of words, such that each pair is rated based on the degree of similarity/relatedness between the words, and then model scores are compared to humans judgements. Another type is analogies test---questions of the form A:B::C:D, meaning A to B is as C to D, and D is to be predicted. While such tests exist for widespread languages, \eg, SimLex-999~\cite{SimLex-999} was translated to four major languages~\cite{leviant2015judgment}, less represented languages suffer from the absence of such resources.  In this work, we attempt to close this gap for Tatar, an agglutinative language with rich morphology, by proposing three evaluation datasets. In general, their use is not limited to embeddings evaluation; they can benefit any system which models semantic/morphosyntactic relationships. For example, they can be used for automated thesauri, dictionary building, machine translation~\cite{he-MT}, or semantic parsing~\cite{Beltagy-sem_pars}.
\section{Related Works}
The most well-known similarity/relatedness datasets for English are RG~\cite{rubenstein1965contextual}, WordSim-353~\cite{Finkelstein}, MEN~\cite{bruni-MEN}, and SimLex-999~\cite{SimLex-999}. The problem with WordSim-353 and MEN is that there's no distinction between similarity and relatedness concepts, and we tried to address it in our work. The analogies task was first introduced by Mikolov \etal~\cite{mikolov2013efficient} and then adapted for a number of languages. The adaptation process is not trivial since analogies should examine specifics of a given language while too much customization would make datasets incomparable. We aimed at finding a compromise between these two extremes.
\section{Proposed Datasets}
\subsection{Similarity Dataset}
For constructing the Similarity dataset we used the WordSim-353 dataset; namely, it's version by Agirre \etal~\cite{agirre-wordsim}, in which they split the original dataset into two subsets, one for similarity, and the other for relatedness evaluation. We took the first subset (for similarity), consisting of a total of 202 words, and manually translated it to Tatar. Then we removed or replaced with analogies when possible pairs containing rare words, and those which needed to be adapted to account for cultural differences, \eg, the Harvard-Yale pair was replaced with \textit{KFU-KAI}, which are acronyms of two largest Tatarstan university names.  We also filtered out most of the pairs with loanwords from Russian.

We defined the distribution of synonymy classes we want to be present in the dataset as follows:

\begin{itemize}
  \item Strong synonyms, $22\%$;
  \item Weak synonyms, $30\%$;
  \item Co-hyponyms, $17\%$;
  \item Hypernym-hyponyms, $15\%$;
  \item Antonyms, $5\%$; and
  \item Unrelated, $11\%$.
\end{itemize}

These categories and percentages were chosen to represent the diversity in synonymy relations and to focus on not-so-obvious pairs, which constitute the majority ($67\%$) of the dataset---everything apart from strong synonyms and unrelated words.
After the described preprocessing, we split the remaining pairs between these categories and added our own pairs such that the total is still 202 and all categories are full.  We used SimLex-999 to find some of the new word pairs. From the part-of-speech point of view, the dataset is mostly build up from nouns, $87\%$, a small fraction of adjectives, $12\%$, and $1\%$ of mixed pairs.

\subsection{Relatedness Dataset}

The Relatedness dataset was constructed using a similar procedure - we took the second relatedness subset of WordSim-353, and translated it adapting/replacing pairs with rare or irrelevant words. For this dataset we kept more loanwords from Russian to keep it close to the original dataset for comparability. The dataset contains 252 words, $98\%$ of which are nouns, and $2\%$ are mixed ones.

\subsubsection{Annotation} 

Here we explain the process of obtaining human scores for the datasets described above. We constructed a survey for each dataset and provided a set of instructions to respondents. For Similarity dataset it was motivated by SimLex-999: we showed the examples of synonyms (life-existence), nearly synonyms (hair-fur), and clearly explained the difference between similarity and relatedness concepts using such examples as car-road. For Relatedness dataset instructions explained different association types (by contrast, by causation, \etc). Then annotators were asked to rate each pair by assigning it to one of the four categories. Depending on the survey, the options were different, below are the versions for \textit{similarity} and (relatedness):

\begin{enumerate}
\item Words are absolutely \textit{dissimilar} (unrelated)
\item Weak \textit{similarity} (relatedness)
\item Moderate \textit{similarity} (relatedness)
\item Words are \textit{very similar or identical} (strongly related)
\end{enumerate}

Later this scale was converted to 0--10 to match the existing datasets (1 $\to$  0; 2~$\to$~\( \frac{10}{3} \); 3 $\to$ \( \frac{20}{3} \); 4 $\to$ 10). A total of 13 respondents rated each dataset, all native Tatar speakers.  Inter-annotator agreement measured as average Spearman's $\rho$ between pairwise ratings equals $0.68$ for Similarity and $0.61$ for Relatedness dataset.

\subsection{Analogies Dataset}
\label{sec:analdescr}
We used several existing analogies datasets to identify common categories to include to our new dataset, namely, the original one~\cite{mikolov2013efficient}, and ones for Czech~\cite{svoboda2016new} and Italian~\cite{berardi2015gotoItaly}. We identified 8 such categories (marked below as~\dag). We also included new categories, most of which explore the morphological richness of the Tatar language, and some account for cultural/geographic characteristics. We applied a frequency threshold: we did not include pairs where any of the words was not in the top 100 000 of most frequent words.
\clearpage
The final list of the categories is as follows: 

\subsubsection{Semantic Categories} ~\\

\noindent\textbf{Capital-country}\textsuperscript{\dag}: capitals of countries worldwide, mostly taken from the original dataset, plus four additional countries, \eg, ``Prague''-``Czech Republic'', the latter expressible as a single word in Tatar. 51 pairs. \\
\textbf{Country-currency}\textsuperscript{\dag}: national currencies worldwide, \eg, ``Turkey''-``lira''. 11 pairs.\\
\textbf{Capital-republic inside Russia}: capitals and names of republics, which are federal subjects of Russia, \eg, ``Kazan''-``Tatarstan''. 14 pairs.\\
\textbf{Man-woman}\textsuperscript{\dag}: family relations, like brother-sister, but also masculine/feminine forms of professions and honorifics. \Eg, \textit{afande-khanym}, which can be translated as ``mister''-``missis''. 27 pairs.\\
\textbf{Antonyms (adjectives)}: \eg ``clean''-``dirty''. Differs from `Opposite' category in the original dataset: words here do not share roots. 50 pairs.\\ 
\textbf{Antonyms (nouns)}: \eg, ``birth''-``death'', roots also differ. Both antonym categories were built using the dictionary of Tatar antonyms\footnote{Safiullina, ISBN 5-94113-178-X.}. 50 pairs.\\
\textbf{Name-occupation}: \eg, ``Tolstoi''-``writer'', for people famous in / associated with Tatarstan Republic. 40 pairs.

\subsubsection{Syntactic Categories} ~\\

\noindent\textbf{Comparative}\textsuperscript{\dag}: positive and comparative forms of adjectives, \eg ``big''-``bigger''. 50 pairs.\\
\textbf{Superlative}\textsuperscript{\dag}: positive and superlative forms of adjectives. Superlatives in Tatar are usually formed by adding separate \textit{in'} ``most'' word before the adjective, but sometimes the first part of the word is repeated twice, \eg, \textit{yashel-yam-yashel}, ``green''-``greenest''. We include 30 such pairs.\\
\textbf{Opposite}\textsuperscript{\dag}: basically antonyms sharing the root, \eg, ``tasty''-``tasteless''. 45 pairs.\\
\textbf{Plural}\textsuperscript{\dag}: singular and plural word forms, \eg, ``school''-``schools''. Contains two subcategories: for nouns, 50 pairs, and pronouns, 10 pairs.\\
\textbf{Cases}: specific for Tatar, which has 6 grammatical cases: nominative, possessive, dative, accusative, ablative, and locative. We pair 30 words in nominative case with their forms in each of other cases, resulting in 5 subcategories. The example of nominative-possessive subcategory can be ``he''-``his''. Words are a mix of nouns and pronouns.\\
\textbf{Derivation (profession)}: nouns and derived profession names, \eg, ``history''-``historian''. 25 pairs.\\
\textbf{Derivation (adjectives)}: nouns and derived adjectives, \eg, ``salt''-``salty''. 30 pairs.

The following 12 categories explore different verb forms. Tatar, an agglutinative language, is extremely rich \wrt word forms.  When it comes to verbs, the form depends on negation, mood, person, number, and tense, among others. We explore only some combinations of these aspects, which we think are most important. To make it easier to comprehend, we only provide examples (for the verb ``go''), and the above-mentioned details can be implied from them, while main aspect is described by category name. We picked 21 verbs and put them in different forms as required by the following categories:\\
\textbf{Negation}: ``he goes''-``he doesn't go''.\\
\textbf{Mood (imperative)}: ``to go''-``go!''.\\
\textbf{Mood (conditional)}: ``go!''-``if he goes''.\\
\textbf{Person1-2}: ``I go''-``you go''.\\
\textbf{Person1-3}: ``I go''-``he goes''.\\
\textbf{Plural, 1 person}: ``I go''-``we go''.\\
\textbf{Plural, 2 person}:``you go (alone)''-``you go (as a group)''.\\
\textbf{Plural, 3 person}: ``he goes''-``they go''.\\
\textbf{Tense, past (definite)}\dag: ``he goes''-``he went''.\\
\textbf{Tense, past (indefinite)}: ``he goes''-``he probably went''.\\
\textbf{Tense, future (definite)}: ``he goes''-``he will go''.\\
\textbf{Tense, future (indefinite)}: ``he goes''-``he will probably go''.\\
\textbf{Verbal adverbs, type 1}: imperative and adverb, \eg ``go!''-``while going''. For the same 21 verbs.\\
\textbf{Verbal adverbs, type 2}: imperative and adverb, \eg ``go!''-``on arrival''. For the same 21 verbs.\\
\textbf{Passive voice}: two verbs, second being a passive voice derivation from the first, \eg, ``he writes''-``it is being written''. 25 pairs.\\

So, in total we constructed 34 categories: 7 semantic and 27 syntactic ones. For each category/subcategory we generated all possible combinations of pairs belonging to it, \eg, the first category (capital-country) contains 51 unique pairs, hence, $50 \cdot 51 = 2550$ combinations. So, we have 10004 semantic and 20140 syntactic questions, in total 30144.

\section{Experiments and Evaluation}
In this section we evaluate three word embedding models, Skip-gram with negative sampling SG~\cite{MikolovDistr}, FastText~\cite{bojanowski2016enriching}, and GloVe~\cite{glove}, with the proposed datasets. These models were chosen for evaluation for their popularity and ease of training. As for Fasttext, it was chosen also because it works with $n$-grams, hence, was expected to handle complex morphology of Tatar better. We trained SG and FastText using gensim\footnote{\url{https://radimrehurek.com/gensim}} library and GloVe using the original code\footnote{\url{https://github.com/stanfordnlp/GloVe}}. Private 126\,M tokens corpus, kindly provided by Corpus of Written Tatar\footnote{\url{http://www.corpus.tatar/en}}, was used to train these models. The corpus was obtained primarily from web-resources and is made up of texts in different genres, such as news, literature, official.

All models were trained with 300~dimensions and window size~5. We used different versions of SG and FastText in our experiments, all trained for 10~epochs using batch size~128. For other parameters, we will refer to minimum word count as $\mathit{mc}$, subsampling threshold as $\mathit{sub}$, negative samples number as $\mathit{neg}$, minimum/maximum $n$-gram length as $\mathit{gram\_l}$. When tuning these parameters we were focused on Analogies task, and for Similarity and Relatedness tasks we chose best results among all trained models. 

As for GloVe, there is only one version for all experiments---trained for 100~epochs with x\_max parameter set to 100, and other parameters set to default.

\subsection{Similarity and Relatedness Results}
For evaluation we calculate Spearman's~$\rho$ correlation between average human score and cosine similarity between embeddings of words in pairs.  We report Spearman's~$\rho$ for SG, FastText, and GloVe for both tasks and parameter configuration which led to the best performance in Table~\ref{rhotable}.
\begin{table}
\captionsetup{justification=centering}
\caption{Spearman's~$\rho$ for Similarity and Relatedness and model parameters.}
\footnotesize
\centering
\begin{tabular}{p{1.8cm}p{4.2cm}p{3.8cm}}
\toprule
\bfseries Model & \bfseries Similarity & \bfseries Relatedness \\
\midrule
SG & 0.52 & 0.60 \\ 
& $\mathit{mc}$=5, $\mathit{sub}$=0, $\mathit{neg}$=64 & $\mathit{mc}$=5, $\mathit{sub}$=0, $\mathit{neg}$=20 \\ 
\midrule
FastText & \textbf{0.54} & \textbf{0.62} \\
&  $\mathit{mc}$=2,~$\mathit{sub}$=$1e^{-4}$,~$\mathit{neg}$=64, \raggedright $\mathit{gram\_l}$=3/6 & $\mathit{mc}$=2,~$\mathit{sub}$=$1e^{-4}$,~$\mathit{neg}$=64, $\mathit{gram\_l}$=3/6 \\ 
\midrule
GloVe & 0.48 & 0.53 \\
\bottomrule
\end{tabular}
\label{rhotable}
\end{table}

As we see, FastText does better on both tasks and GloVe performs substantially worse, while for all models the similarity task appears to be trickier, which does not positively correlate with the inter-annotator agreement.

\subsection{Analogies Results}
\label{sec:analresults}
We follow the same evaluation procedure as in the original work~\cite{mikolov2013efficient}, so, we report accuracy, where true prediction means exact match (1\textsuperscript{st} nearest neighbor). We used same parameters for both SG and FastText: $\mathit{mc}=5$, $\mathit{sub}=0$, $\mathit{neg}=20$, initial/minimal learning rates were set to $0.05$ and $1e^{-3}$ respectively. For FastText $\mathit{gram\_l}$ was set to $3/8$. The results are presented in Table~\ref{acctable}, where we compare performance of SG, FastText, and GloVe for each category of analogies. For summaries (bold lines), we calculate average over categories instead of overall accuracy, to cope with imbalance in categories' sizes: this way all categories have the same impact on summary score. 
\begin{table}
\caption{Accuracy (\%) per analogy category.}
\centering
\renewcommand{\arraystretch}{1.15}
\sisetup{table-format=2.2}  
\rowcolors{3}{white}{gray!10}
\footnotesize
\begin{tabular}{p{4cm}SSS}
\toprule
\bfseries Category & {\bfseries SG} & {\bfseries FastText} &  {\bfseries GloVe} \\
\midrule
\multicolumn{4}{l}{\bfseries Semantic categories} \\ 
\midrule
capital-country	&	40.51	&	32.31	&	15.53	\\
country-currency	&	4.55	&	5.45	&	5.45	\\
capital-republic-rf	&	33.52	&	23.08	&	30.22	\\
man-woman	&	40.46	&	38.32	&	41.03	\\
adj-antonym	&	8.61	&	7.43	&	6.73	\\
noun-antonym	&	8.78	&	7.39	&	7.67	\\
name-occupation	&	27.95	&	12.76	&	15.71 \setcounter{rownum}{0} \\ \midrule
\bfseries Semantic average	&\bfseries 23.48 &\bfseries 18.11 &\bfseries 17.48  \\ \midrule
\multicolumn{4}{l}{\bfseries Syntactic categories} \\ \midrule
comparative	&	76.04	&	76.78	&	44.65	\\
superlative	&	26.90	&	33.33	&	5.29	\\
opposite	&	15.15	&	13.08	&	5.30	\\
plural-nouns	&	38.04	&	41.59	&	16.94	\\
plural-pronouns	&	23.33	&	27.78	&	16.67	\\
cases-possessive	&	32.07	&	39.66	&	5.40	\\
cases-dative	&	14.83	&	12.53	&	3.33	\\
cases-accusative	&	30.80	&	36.09	&	7.70	\\
cases-ablative	&	5.98	&	12.64	&	1.15	\\
cases-locative	&	12.99	&	15.17	&	4.94	\\
profession	&	20.33	&	17.33	&	8.50	\\
noun-adj	&	7.01	&	4.14	&	6.09	\\
negation	&	52.14	&	57.62	&	19.76	\\
imperative	&	40.24	&	44.52	&	8.33	\\
conditional	&	40.00	&	52.14	&	7.14	\\
person1-2	&	65.71	&	81.90	&	28.57	\\
person1-3	&	74.29	&	71.67	&	64.52	\\
plural-1person	&	91.43	&	93.57	&	65.71	\\
plural-2person	&	63.57	&	95.24	&	35.24	\\
plural-3person	&	91.90	&	86.43	&	75.24	\\
present-past-def	&	82.62	&	88.33	&	65.48	\\
present-past-indef	&	87.86	&	86.43	&	84.29	\\
present-future-def	&	45.24	&	70.24	&	26.90	\\
present-future-indef	&	55.95	&	55.71	&	28.81	\\
verbal-adv-1	&	23.81	&	29.05	&	9.76	\\
verbal-adv-2	&	40.24	&	50.71	&	5.48	\\
passive-voice	&	34.50	&	35.33	&	14.67	\setcounter{rownum}{0} \\ \midrule
\bfseries Syntactic average	&\bfseries 44.18 &\bfseries 49.22 &\bfseries 24.66	 \\ \midrule
\bfseries All average	&\bfseries 39.92 &\bfseries  42.82 &\bfseries 23.18	 \\ 
\bottomrule
\end{tabular}
\label{acctable}
\end{table}

We observe that syntactic questions are substantially easier to answer for all models, which indicates they better capture morphological relationships than semantic ones. If we examine syntactic categories, we see that analogy questions with nouns and adjectives (from the top to \textit{noun-adj}) are on average more challenging for models to answer than questions with verbs (from \textit{negation} till the end). Let's take, for example, plural categories for nouns and verbs---the difference is drastic, accuracy for plural verb categories is 2-4 times higher than that for nouns. This suggests that more emphasis should be made on modeling grammatical aspects of nouns and adjectives in the future. 

Overall, categories' complexity varies a lot, \eg, all models show more than $80\%$ accuracy in \textit{present-past-indefinite} category, whereas \textit{country-currency} and \textit{noun-adj} are especially complex. Among other categories which challenge our models we see some types of noun cases, as well as antonyms: whether sharing the root (\textit{opposite}) or not (\textit{adj-antonym}). 

As expected, SG and FastText exhibit more similar behaviour, when compared to GloVe, due to their architectural commonalities.
GloVe performs much worse in syntactic categories; for most of them the results are incomparably lower. We see that SG beats FastText in semantic questions, but is inferior to it in syntactic ones, which is as expected because FastText by construction should learn more about morphology, and Tatar is morphologically rich language.

\subsection{Comparison with English}

To make a comparison between languages, we took a 126\,M tokens snippet of English News Crawl 2017 corpus\footnote{\url{http://www.statmt.org/wmt18/translation-task.html}}, trained models with same parameters as described in section~\ref{sec:analresults}, and evaluated them against the original English datasets.   

\subsubsection{Similarity and Relatedness}
First, we measured Spearman's $\rho$ for similarity and relatedness splits of WordSim-353 by Agirre \etal~\cite{agirre-wordsim} for trained English models, and compared with corresponding results for Tatar in Table~\ref{rho_entable}. 

\begin{table}
\caption{Spearman's~$\rho$ for Similarity and Relatedness for Tatar and English.}
\footnotesize
\centering
\begin{tabular}{p{2cm}p{2cm}p{1.5cm}p{1.3cm}}
\toprule
\bfseries Model & \bfseries Task & \bfseries Tatar & \bfseries English \\
\midrule
\multirow{2}{*}{SG} & Similarity & 0.52 & \textbf{0.68} \\ 
& Relatedness & 0.60 & \textbf{0.55} \\ \midrule
\multirow{2}{*}{FastText} & Similarity & \textbf{0.54} & \textbf{0.68} \\
& Relatedness &  \textbf{0.62} & \textbf{0.55} \\ \midrule
\multirow{2}{*}{GloVe} & Similarity &  0.48 & 0.46 \\
& Relatedness &  0.53 & 0.39 \\
\bottomrule
\end{tabular}
\label{rho_entable}
\end{table}
For the English language, performance of SG and FastText on similarity and relatedness tasks is the same, and for GloVe the numbers are lower. If we compare with Tatar, we see that the values lie in the same range which is good because this indicates that the datasets are probably comparable, as it was designed. Interestingly, though, in contrast to Tatar, relatedness task appears to be more difficult for English models.  

\subsubsection{Analogies}
Second, we evaluated trained English models on the original analogies dataset. We compare average accuracies in semantic and syntactic categories for two languages in Table~\ref{entable}.

\begin{table}
\caption{Comparison of average accuracies (\%) for Tatar and English.}
\footnotesize
\centering
\begin{tabular}{p{2cm}p{2cm}p{1.5cm}p{1.3cm}}
\toprule
\bfseries Model && \bfseries Tatar & \bfseries English \\
\midrule
\multirow{2}{*}{SG} & Semantic & \textbf{23.48} & \textbf{53.11} \\ 
& Syntactic & 44.18 & 56.39 \\ \midrule
\multirow{2}{*}{FastText} & Semantic & 18.11 & 41.27 \\  
& Syntactic & \textbf{49.22} & \textbf{65.48} \\ \midrule
\multirow{2}{*}{GloVe} & Semantic & 17.48 & 47.47 \\  
& Syntactic & 24.66 & 36.41 \\
\bottomrule
\end{tabular}
\label{entable}
\end{table}

English FastText falls behind English SG in semantic questions ($41\%$ \vs $53\%$) but outperforms in syntactic questions ($65\%$ \vs $56\%$), the same tendency as it was observed for the Tatar models and in the original work~\cite{bojanowski2016enriching}. Notice that regardless of language GloVe shows poor results in syntactic categories.

What's most interesting, however, is that English models perform much better: they are from $12\%$ to $30\%$ more accurate than their Tatar counterparts. We think that the overall better performance of English models can be partially explained by easier questions: our dataset includes such nontrivial categories as antonyms, cases, and derivations.

\begin{table} 
\caption{Accuracy (\%) of FastText for common categories.}
\centering
\renewcommand{\arraystretch}{1.15}
\sisetup{table-format=2.2}  
\rowcolors{3}{white}{gray!10}
\footnotesize
\begin{tabular}{p{4cm}SS}
\toprule
\bfseries Category & {\bfseries Tatar} & {\bfseries English} \\
\midrule
\multicolumn{3}{l}{\bfseries Semantic categories} \\ 
\midrule
country-capital	&	32.31	&	62.04	\\
country-currency	&	5.45	&	4.02    \setcounter{rownum}{0}	\\ 
man-woman	&	38.32	&	52.37	\\ \midrule
\multicolumn{3}{l}{\bfseries Syntactic categories} \\ \midrule
comparative	&	76.78	&	86.94	\\
superlative	&	33.33	&	75.58	\\
opposite	&	13.08	&	35.71	\\
plural-nouns	&	41.59	&	76.43	\\
present-past-def    &	88.33	&	56.60	\\
\bottomrule
\end{tabular}
\label{tatengtable}
\end{table}

To analyze it further, we can compare results for Tatar and English category-wise. We mentioned in section~\ref{sec:analdescr} that 8 common categories were selected to be included in Analogies dataset to make it comparable with existing ones. Now in Table~\ref{tatengtable} we report accuracy values for these common categories for Tatar and English languages tested on FastText model, as it showed better overall results for both languages.

We see from the results that there's some correlation between Tatar and English models. Both models perform well in comparatives, worse in opposites, and very poorly in \textit{country-currency} category. Surprisingly, Tatar model performs better in past tense. For other categories gaps are too huge to compare. For \textit{country-capital} the better performance may be explained by the fact that the news dataset used for training English models is probably an especially good resource for such data compared to Tatar corpus. For superlatives the probable reason is the earlier explained characteristic---superlatives in Tatar are usually formed by adding a separate word, and adjectives included in this category represent a very small subset of adjectives, which have one-word superlative form.

Overall, the difference may also be due to dissimilarities of training corpora. There are no large parallel corpora for English and Tatar, unfortunately; if this was the case, the comparison would be more accurate.

\section{Conclusion}

In this paper, we introduced three new datasets for evaluating and exploring word embeddings for the Tatar language. The datasets were constructed accounting for cultural specificities and allow in-depth analysis of embedding performance \wrt various language characteristics. We examined the performance of SG, FastText and GloVe word embedding models on introduced datasets, showing that for all three tests there is much room for improvement. Cross-language comparison demonstrated that models' performance varies greatly with language; nonetheless, similar trends were observed across languages.

\section{Acknowledgments}
We thank Mansur Saykhunov, the main author and maintainer of Corpus of Written Tatar\footnote{\url{http://www.corpus.tatar/en}}, for providing us data; and all respondents of the surveys for constructing Similarity/Relatedness datasets. This research was partially supported by the Brazilian National Council for Scientific and Technological Development (CNPq grant \#~307425/2017-7).

\bibliographystyle{splncs04}
\bibliography{my_bib}

\end{document}